\documentclass{article}

\usepackage[preprint]{neurips_2024}

\usepackage{graphicx} 
\usepackage{amsmath}

\usepackage[utf8]{inputenc}
\usepackage{CJKutf8}

\usepackage[utf8]{inputenc} 
\usepackage[T1]{fontenc}    
\usepackage{hyperref}       
\usepackage{url}            
\usepackage{booktabs}       
\usepackage{amsfonts}       
\usepackage{nicefrac}       
\usepackage{microtype}      
\usepackage{xcolor}         

\title{EEG2TEXT-CN: \\ An Exploratory Study of Open-Vocabulary Chinese Text-EEG Alignment via Large Language Model and Contrastive Learning on ChineseEEG}

\author{%
  Jacky Tai-Yu Lu\thanks{Equal contribution}\quad\textsuperscript{1}, 
  Jung Chiang\footnotemark[1]\quad\textsuperscript{1,2}, 
  Chi-Sheng Chen\footnotemark[1]\quad\textsuperscript{3},\\
  \textbf{Anna Nai-Yun Tung\textsuperscript{1,4},}
  \textbf{Hsiang Wei Hu\textsuperscript{1},}
  \textbf{Yuan Chiao Cheng\textsuperscript{1,6}}\\
  \textsuperscript{1}Department of Artificial Intelligence in Healthcare,\\ International Academia of Biomedical Innovation Technology, Reno, USA  \\
  \textsuperscript{2}National Taiwan University  \\
  \textsuperscript{3}Neuro Industry Research, Neuro Industry, Inc.  \\
  \textsuperscript{4}Department of Biomedical Engineering, \\University of Southern California \\
  \textsuperscript{5}Taiwan Artificial Intelligence Association \\
  \texttt{b0090100@gmail.com, Lloyd4127@icloud.com,m50816m50816@gmail.com,} \\
  \texttt{naiyun@usc.edu, Hw.Hsiang.wei@gmail.com, yuanchiao.cheng@gmail.com}
}

\begin{document}

\maketitle

\begin{abstract}
We propose EEG2TEXT-CN, which, to the best of our knowledge, represents one of the earliest open-vocabulary EEG-to-text generation frameworks tailored for Chinese. Built on a biologically grounded EEG encoder (NICE-EEG) and a compact pretrained language model (MiniLM), our architecture aligns multichannel brain signals with natural language representations via masked pretraining and contrastive learning. Using a subset of the ChineseEEG dataset, where each sentence contains approximately ten Chinese characters aligned with 128-channel EEG recorded at 256 Hz, we segment EEG into per-character embeddings and predict full sentences in a zero-shot setting. The decoder is trained with teacher forcing and padding masks to accommodate variable-length sequences. Evaluation on over 1,500 training-validation sentences and 300 held-out test samples shows promising lexical alignment, with a best BLEU-1 score of 6.38\%. While syntactic fluency remains a challenge, our findings demonstrate the feasibility of non-phonetic, cross-modal language decoding from EEG. This work opens a new direction in multilingual brain-to-text research and lays the foundation for future cognitive-language interfaces in Chinese.
\end{abstract}

\section{Introduction}

Translating brain activity into natural language has become a key ambition across the fields of brain-computer interfaces (BCIs), cognitive neuroscience, and neural language modeling. Among various neuroimaging modalities, electroencephalography (EEG) remains a compelling choice due to its non-invasive nature, fine-grained temporal resolution, and feasibility for real-world deployment. However, most prior EEG-based decoding systems have been constrained by narrow vocabularies, stimulus-locked paradigms, and shallow modeling pipelines. These limitations restrict their ability to capture the continuous and context-rich nature of language processing in the brain.

Recent advances in deep learning and self-supervised representation learning have introduced the possibility of open-vocabulary EEG-to-text generation, where neural activity can be translated into free-form, coherent textual output. This evolution marks a paradigm shift—from classifying discrete EEG events to learning contextualized brain-to-language mappings.

\subsection{Literature Review}

Pioneering work in EEG-to-text decoding includes the EEG2TEXT framework, which introduced a masked EEG modeling strategy and a multi-view transformer architecture to generate structured text from sentence-level EEG signals \cite{liu2024eeg2text}. This method departed from stimulus-aligned decoding by directly learning semantic mappings from raw EEG segments across brain regions. Additionally, DeWave applied quantized variational encoding to derive discrete codex representations of EEG, improving the model’s robustness to individual differences and session noise \cite{duan2023dewave}. These advances highlight the role of unsupervised and contrastive learning in extracting meaningful latent structure from high-dimensional EEG data.

In a complementary direction, \cite{wang2022open} explored zero-shot EEG-based sentiment classification by aligning EEG features with pre-trained language models, demonstrating the potential of semantic alignment in EEG-driven NLP tasks. Meanwhile, the NICE-EEG framework advanced visual decoding from EEG by incorporating spatial and graph attention mechanisms, offering a biologically informed encoding of spatiotemporal neural dynamics \cite{song2023decoding}. NICE-EEG’s use of contrastive learning with semantic embeddings provided a generalizable mechanism for decoding across subjects and sessions.

However, most of these studies are focused on English datasets and alphabetic language structures. The recently published ChineseEEG dataset \cite{mou2024chineseeeg} provides a unique opportunity to investigate EEG decoding in logographic, character-based language contexts, which involve distinct cognitive and neural encoding processes. Despite its richness, no prior model has fully leveraged both advanced decoding architectures and this dataset’s potential to support Chinese sentence-level neural decoding.

\subsection{EEG2TEXT-CN: Multiview Semantic Alignment for Chinese EEG}

In this study, we introduce EEG2TEXT-CN, a unified framework that integrates the biologically grounded encoding of NICE-EEG with the semantic representation capabilities of Chinese pre-trained language models. Our approach is the first to explore open-vocabulary Chinese sentence alignment from EEG, using full-sentence data recorded from native speakers reading two novels in the ChineseEEG corpus.

The proposed architecture consists of two key components:
\begin{itemize}
    \item A spatially aware EEG encoder adapted from NICE-EEG, integrating self-attention and graph attention layers to resolve EEG topography;
    \item A masked pretraining module inspired by EEG2TEXT, enabling the model to learn semantic associations from corrupted EEG-language pairs in a vision-free setting.
\end{itemize}

To enhance alignment, we incorporate pre-trained Chinese language models (we use all-MiniLM-L12-v2 \cite{reimers-2021-sentence-bert} in this work) as semantic anchors for EEG-aligned embeddings. This multimodal fusion enables our model to map sparse EEG signals to meaningful language representations in a zero-shot setting.

\section{Related Work}
While recent advances have led to a surge in EEG-to-image alignment research—such as NICE \cite{song2023decoding}, MUSE \cite{chen2024mind}, QMCL \cite{chen2024quantum}, NERV \cite{chen2024necomimi}, and ATM \cite{li2024visual} relatively few studies have addressed the more challenging task of aligning EEG with natural language. Compared to image stimuli, which evoke strong visual cortex responses, language stimuli engage more abstract, temporally extended neural processes, making the decoding problem more complex. As such, EEG-to-text alignment remains underexplored, particularly in the context of open-vocabulary sentence generation.

EEG-to-text is an emerging area at the intersection of neuroscience and artificial intelligence that aims to decode natural language directly from brain signals. Fueled by advances in deep neural networks and improved EEG recording techniques, this field has evolved from closed-vocabulary recognition to more open-ended, generative language models. However, most research to date has focused exclusively on English, leaving a significant gap in multilingual EEG decoding. In this section, we review key developments in EEG-to-text systems, with a focus on model design, learning paradigms, and available datasets, and we identify the unmet challenges that motivate our study.

\subsection{Closed- vs. Open-Vocabulary EEG Decoding}

Initial studies on EEG-to-text focused on closed-vocabulary tasks with limited semantic space. Recent efforts have shifted toward open-vocabulary generation, where models attempt to produce free-form text from EEG input. For instance, EEG2TEXT~\cite{liu2024eeg2text} introduced a multi-perspective transformer pretrained on EEG data, achieving a 5\% improvement in BLEU and ROUGE scores over prior baselines. These results underscore the potential of combining EEG feature extraction with large language models (LLMs). Nevertheless, the effectiveness of such open-vocabulary systems across diverse languages remains underexplored.

\subsection{Contrastive Learning and Autoencoder-Based Methods}

To bridge the modality gap between EEG signals and text, several studies have leveraged contrastive learning and autoencoding strategies. CET-MAE and E2T-PTR~\cite{wang2024survey} jointly embed EEG and text using multi-stream architectures. E2T-PTR employs a BART-based decoder and shows significant gains (8.34\% ROUGE-1 F1 and 32.21\% BLEU-4) on the ZuCo dataset. While these approaches improve alignment and generation quality, they typically assume well-segmented, English-centric inputs and are rarely validated across languages or writing systems.

\subsection{Imagined Speech Decoding and Silent Communication}

Another important research direction involves decoding imagined or silent speech from EEG. Xiong et al.~\cite{xiong2025synthesizing} proposed ETS, an end-to-end EEG-to-speech framework that generates intelligible speech from imagined input using an X-shaped DNN. The model achieved 91.23\% average accuracy across 13 participants. While promising for silent communication, such models often rely on phoneme-level alignment or audio-based supervision, which are less transferable to ideographic languages like Chinese.

\subsection{Self-Supervised and Discrete Encoding Approaches}

To enhance generalization and data efficiency, self-supervised methods have been proposed. BELT~\cite{zhou2024belt, zhou2024belt2} introduced a D-Conformer that maps EEG into discrete representations, guiding alignment with textual sequences. DeWave~\cite{duan2023dewave} combines discrete variational encoding with pretrained language models to tackle token mismatch and subject variability. These works improve robustness but still primarily target English word-level decoding and lack evaluation in logographic languages.

\subsection{Datasets and Language Limitations}

The field currently relies heavily on the ZuCo dataset~\cite{hollenstein2018zuco}, which contains EEG and eye-tracking data from English readers. Though rich and well-annotated, ZuCo only supports English stimuli. To date, no large-scale public EEG dataset has focused on Chinese, despite its distinct linguistic and visual properties. The absence of such resources poses challenges for developing multilingual brain-computer interfaces.

\subsection{Speech Synthesis and Multimodal Systems}

Other efforts aim to reconstruct speech directly from EEG signals. BTS~\cite{lee2023towards} generates a user's own voice using EEG and phoneme-aware speech models. Similarly, phoneme-level BCIs have been explored using low-cost devices~\cite{larocco2023evaluation}, achieving up to 97\% accuracy. These studies focus more on low-level speech units than on full-text generation, and their dependence on audio corpora limits their applicability to non-phonetic writing systems.

\subsection{Toward Multilingual EEG Decoding}

Collectively, these studies highlight remarkable progress in decoding natural language from EEG. However, nearly all prior work has been confined to alphabetic languages, especially English. The challenges of processing logographic languages like Chinese—where character-level encoding, stroke order, and syntax differ significantly—have not been addressed. To our knowledge, no existing work has explored open-vocabulary Chinese sentence generation from EEG.

In this work, we present the first EEG-to-text model for Chinese. Leveraging the newly released ChineseEEG corpus, we introduce a biologically grounded EEG encoder and integrate it with a language model for sentence-level generation. Our work extends the frontiers of EEG decoding into the multilingual domain, laying the foundation for future studies in Chinese semantic decoding and brain-to-text communication.

\begin{figure}
    \centering
    \includegraphics[width=1\linewidth]{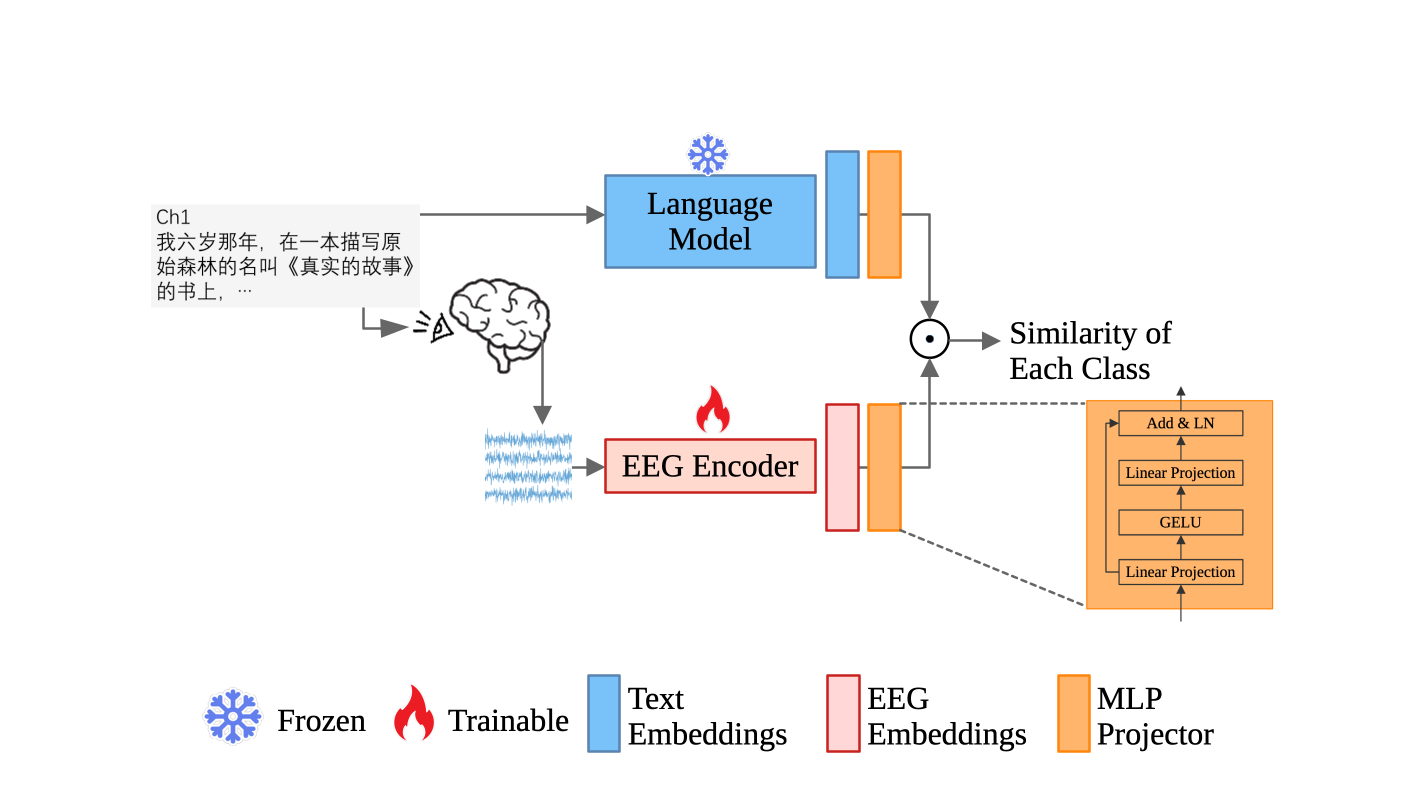}
    \caption{Overview of our EEG2TEXT-CN alignment model. The model takes synchronized input pairs of Chinese text and corresponding EEG signals. The language model (frozen, blue) encodes character-level text into contextual embeddings, while the EEG encoder (trainable, red) maps PCA-compressed EEG segments into the same embedding space. Both embeddings are projected via a shared MLP head and compared using cosine similarity to compute class-level alignment. The model is trained with a contrastive objective.}
    \label{fig:eeg2txtcn}
\end{figure}

\section{Methods}

\subsection{Dataset}

We utilized the ChineseEEG dataset~\cite{mou2024chineseeeg}, a high-density EEG corpus designed for semantic alignment and neural decoding in naturalistic Chinese reading tasks. The dataset contains over 13 hours of EEG and eye-tracking recordings from 10 participants, each silently reading two Chinese novels: \emph{The Little Prince} and \emph{Garnett Dream}. Each character was sequentially highlighted for 350 ms without punctuation, enabling precise alignment between text and EEG. EEG was recorded with a 128-channel EGI Geodesic Sensor Net system at 256 Hz sampling rate.

In our exploratory study, we selected data from sub-04 to sub-09 while they read Chapters 1--18 of \emph{Garnett Dream}, resulting in a total of 77,815 Chinese characters. Each EEG-text pair corresponds to a sentence of up to 10 characters. Based on the fixed character duration (0.35 s), we segmented EEG into 90 time points per character (256 $\times$ 0.35 $\approx$ 89.6, rounded to 90), yielding a tensor of shape (number of characters, 128, 90) for each sentence.

To reduce computational cost, we applied principal component analysis (PCA) \cite{elhaik2022principal} along the channel dimension, compressing each sample to shape (number of characters, 90, 8). During training, sentences of different lengths were zero-padded to a maximum of 10 characters. A \texttt{padding mask} was used during feedforward to prevent gradient updates on padded tokens. However, this padding design makes prediction more difficult, especially when sentence length is unknown during inference.

\begin{figure}
    \centering
    \includegraphics[width=1\linewidth]{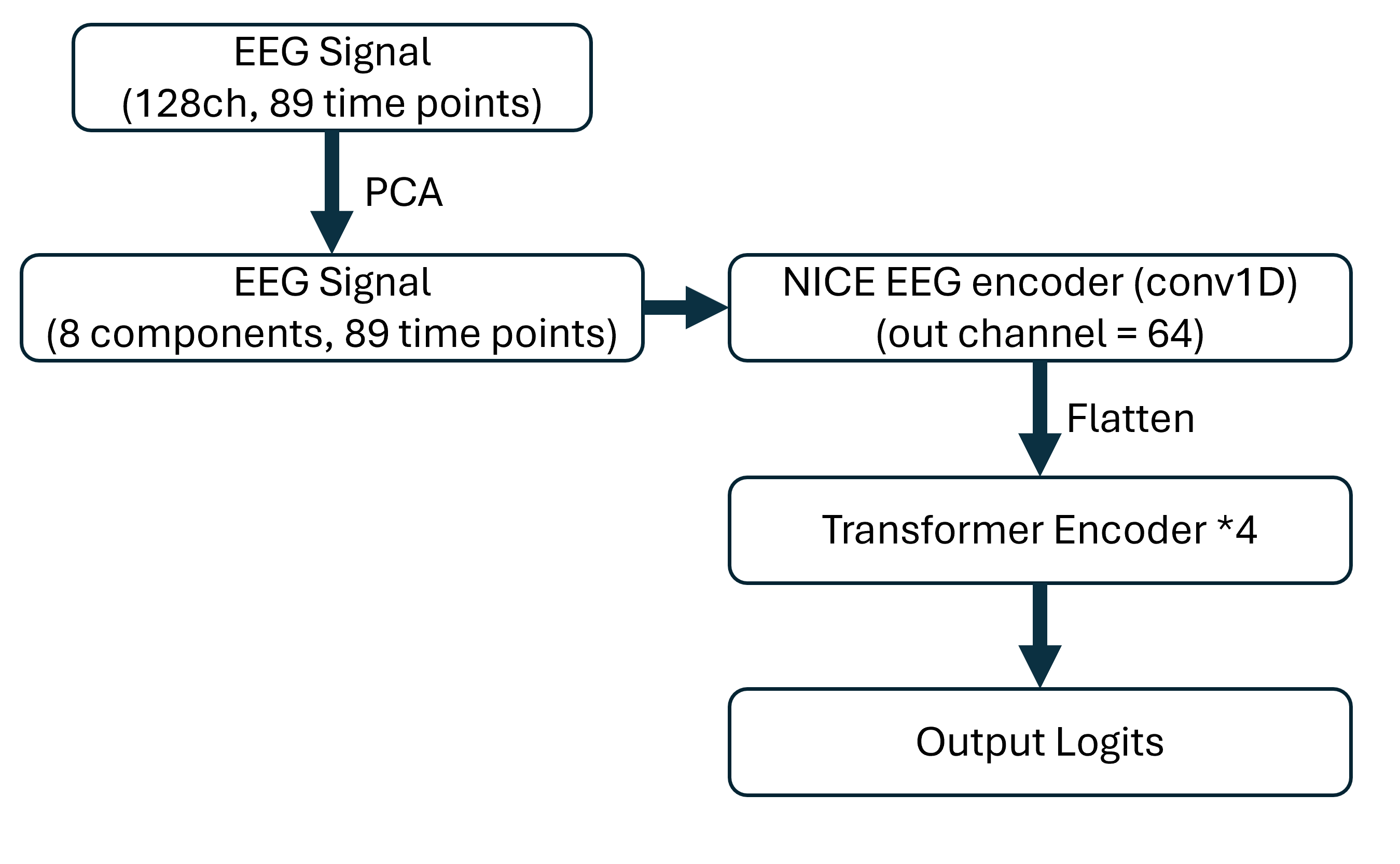}
    \caption{EEG encoder-decoder Pipeline.
The pipeline begins with a raw EEG signal of shape (128 channels, 89 time points), which is reduced to 8 principal components via PCA. The compressed EEG data is then passed through a NICE EEG encoder using a 1D convolutional layer with 64 output channels. The resulting features are flattened and processed through a stack of 4 Transformer encoder layers. Finally, the output is projected to the target class logits.}
    \label{fig:nice}
\end{figure}

\subsection{Model Architecture}

We adopted the convolutional backbone from NICE-EEG \cite{song2023decoding} as our EEG encoder, due to its proven performance in EEG-image alignment tasks. This architecture effectively captures spatial and temporal dynamics in high-density EEG data. The details is shown in Figure~\ref{fig:nice}.

For the language encoder, we used the lightweight transformer model all-MiniLM-L12-v2 from Hugging Face, which offers efficient contextual embeddings. Due to computational constraints, we adopted a character-level input strategy. Although the tokenizer may merge frequent character sequences (e.g., \begin{CJK}{UTF8}{bsmi}“全世界”\end{CJK}) into a single token, we bypass such ambiguity by enforcing per-character input and output, rather than relying on learned subword boundaries.

Decoding is performed in an autoregressive fashion, assuming known sentence start and fixed-length input.

\subsection{Model Formulation}

We formulate EEG2TEXT-CN as a sequence-to-sequence encoder-decoder model that generates Chinese text from EEG signals. Each sentence consists of up to \( N \) characters, with each character associated with an EEG segment \( \mathbf{X}_i \in \mathbb{R}^{T \times C} \), where \( T = 89 \) time points (corresponding to 350 ms at 256 Hz) and \( C = 8 \) spatial components obtained through PCA. The full input for one sentence is represented as \( \mathbf{X} = [\mathbf{X}_1, \dots, \mathbf{X}_N] \in \mathbb{R}^{N \times T \times C} \). The details is shown in Figure~\ref{fig:eeg2txtcn}.

To encode the EEG segments, we use a convolutional encoder \( f_{\text{enc}} \) based on the NICE-EEG architecture. Each segment \( \mathbf{X}_i \) is transformed into a fixed-length vector \( \mathbf{h}_i = f_{\text{enc}}(\mathbf{X}_i) \in \mathbb{R}^{d} \), resulting in an encoded sequence \( \mathbf{H} = [\mathbf{h}_1, \dots, \mathbf{h}_N] \in \mathbb{R}^{N \times d} \). Positional encodings are added to \( \mathbf{H} \) before passing it to the decoder.

The decoder is a Transformer conditioned on both the encoded EEG sequence and previously generated tokens. During training, we apply teacher forcing: the decoder receives the shifted ground truth token sequence \( \mathbf{Y}_{<t} \) and outputs predicted logits for the next token:
\begin{equation}
\hat{y}_t = \text{Softmax}(W_o \cdot \hat{\mathbf{y}}_t + b_o),
\end{equation}
where \( W_o \in \mathbb{R}^{|V| \times d} \) and \( |V| = 250002 \) denotes the vocabulary size including [BOS] and [EOS] tokens.

We optimize the model using cross-entropy loss over the target sequence:
\begin{equation}
\mathcal{L} = -\sum_{t=1}^{N} \log p(y_t \mid y_{<t}, \mathbf{X}).
\end{equation}
Padding masks are applied to ensure loss is only computed over valid character positions. This framework allows autoregressive generation of Chinese text from EEG signals by leveraging both spatial and temporal dynamics of the brainwave data.

\subsection{Training and Evaluation}
We constructed our training and validation sets from the first 100 sentence segments of 3 chapters of \emph{Garnett Dream}, across five participants. The model is trained for 50 epochs. This resulted in 1,500 total samples, randomly split into 1,200 for training and 300 for validation. We employed teacher forcing \cite{williams1989learning} during training, feeding ground-truth characters step-by-step into the decoder to stabilize learning and improve convergence.

For evaluation, we selected the sixth participant (sub-06), extracting the same 3 chapters and 100 segments per chapter, yielding a test set of 300 samples. This subject-wise split ensures generalization to unseen subjects and avoids data leakage.

We evaluate EEG2TEXT-CN on the ChineseEEG dataset by analyzing the output probability distribution over individual characters for each predicted sentence. To enable a fair and interpretable comparison, we selected the best-performing model checkpoint and computed BLEU scores (BLEU-1 to BLEU-4) on the top-1 predictions.

BLEU (Bilingual Evaluation Understudy) \cite{papineni2002bleu} measures the n-gram precision between the predicted and ground truth sequences, with a brevity penalty to penalize overly short outputs. The BLEU score is computed as:

\begin{equation}
\text{BLEU} = \text{BP} \cdot \exp\left( \sum_{n=1}^{N} w_n \log p_n \right)
\end{equation}

Here, \( p_n \) is the modified n-gram precision up to order \( N \), \( w_n \) is the weight (typically uniform), and \( \text{BP} \) is the brevity penalty:

\begin{equation}
\text{BP} = 
\begin{cases}
1 & \text{if } c > r \\
\exp(1 - r/c) & \text{if } c \leq r
\end{cases}
\end{equation}

where \( c \) is the length of the candidate (predicted) sentence, and \( r \) is the length of the reference sentence.

\medskip

Specifically:
\begin{itemize}
    \item \textbf{BLEU-1:} considers unigram (single-character) overlap;
    \item \textbf{BLEU-2:} includes bigram (two-character sequence) precision;
    \item \textbf{BLEU-3:} accounts for trigrams;
    \item \textbf{BLEU-4:} evaluates fluency with four-character sequences.
\end{itemize}

This fine-grained evaluation reflects the model’s ability to capture both lexical accuracy and short-range fluency. Despite the open-vocabulary and low-resource setting, our model demonstrates promising character-level alignment between EEG and text. These results suggest that our framework provides a viable foundation for cross-modal brain-to-language modeling in Chinese, opening new directions for future multilingual EEG-to-text research.

\begin{table}[h]
\centering
\caption{BLEU-n scores on the test set using the best-performing model (Epoch 49).}
\begin{tabular}{lcc}
\toprule
\textbf{Metric} & \textbf{n-gram Level} & \textbf{Score} \\
\midrule
BLEU-1 & Unigram         & 0.0638 \\
BLEU-2 & Bigram          & 0.0212 \\
BLEU-3 & Trigram         & 0.0152 \\
BLEU-4 & 4-gram          & 0.0132 \\
\bottomrule
\end{tabular}
\label{tab:bleu_results}
\end{table}

\begin{table}[h]
\centering
\caption{Comparison between Ground Truth and Predicted Results with BLEU scores. The first result is from epoch 6 and the second result is from epoch 49.}
\begin{tabular}{p{4.5cm} p{4.5cm} cccc}
\toprule
\textbf{Ground Truth (GT)} & \textbf{Predicted Result (PR)} & \textbf{BLEU-1} & \textbf{BLEU-2} & \textbf{BLEU-3} & \textbf{BLEU-4} \\
\midrule
\begin{CJK}{UTF8}{bsmi}全世界的狼都有一个共\end{CJK} & \begin{CJK}{UTF8}{bsmi}在草原东北端一块马蹄甲\end{CJK} & 0.0909 & 0.0302 & 0.0216 & 0.0189 \\
\begin{CJK}{UTF8}{bsmi}全世界的狼都有一个共\end{CJK} & \begin{CJK}{UTF8}{bsmi}石和一块块岩。这样奔跑\end{CJK} & 0.0909 & 0.0302 & 0.0216 & 0.0189 \\
\begin{CJK}{UTF8}{bsmi}同的习性，在严寒的冬天\end{CJK} & \begin{CJK}{UTF8}{bsmi}石。终于草棚上的猎人和\end{CJK} & 0.0909 & 0.0302 & 0.0216 & 0.0189 \\
\begin{CJK}{UTF8}{bsmi}同的习性，在严寒的冬天\end{CJK} & \begin{CJK}{UTF8}{bsmi}在草原东北端一块马蹄甲\end{CJK} & 0.0909 & 0.0302 & 0.0216 & 0.0189 \\
\begin{CJK}{UTF8}{bsmi}集合成群，平时单身独处。\end{CJK} & \begin{CJK}{UTF8}{bsmi}石。终于草棚上的猎人和\end{CJK} & 0.0830 & 0.0275 & 0.0197 & 0.0172 \\
\begin{CJK}{UTF8}{bsmi}集合成群，平时单身独处。\end{CJK} & \begin{CJK}{UTF8}{bsmi}在草原东北端一块马蹄甲\end{CJK} &  0.0000 & 0.0000 & 0.0000 & 0.0000 \\
\begin{CJK}{UTF8}{bsmi}眼下正是桃红柳绿的春\end{CJK} & \begin{CJK}{UTF8}{bsmi}石。终于草棚上的猎人和\end{CJK} & 0.0909 & 0.0302 & 0.0216 & 0.0189 \\
\begin{CJK}{UTF8}{bsmi}眼下正是桃红柳绿的春\end{CJK} & \begin{CJK}{UTF8}{bsmi}在草原东北端一块马蹄甲\end{CJK} & 0.0000 & 0.0000 & 0.0000 & 0.0000 \\
\bottomrule
\end{tabular}
\label{tab:gt_pr_bleu}
\end{table}

\section{Results Analysis}
We evaluated the EEG2TEXT-CN model using BLEU-1 through BLEU-4 scores on the ChineseEEG test set, the results are in Table~\ref{tab:bleu_results} and Table~\ref{tab:gt_pr_bleu}. The best-performing checkpoint (Epoch 49) yielded the following results: BLEU-1 = 0.0638, BLEU-2 = 0.0212, BLEU-3 = 0.0152, and BLEU-4 = 0.0132. These scores indicate that the model demonstrates better performance at the character level, while struggling with longer n-gram coherence and syntactic fluency.

To better understand these results, we qualitatively analyzed predicted sentences against ground truth (GT) sentences. In a few cases, the model successfully captured overlapping characters or semantic fragments from the GT (e.g., phrases such as \begin{CJK}{UTF8}{bsmi}“石。于草棚上的人和”\end{CJK} partially match GT sentences), which leads to non-zero BLEU scores. However, in most instances, the predictions were only loosely related to the GT and often defaulted to high-frequency tokens in the training corpus, indicating a strong lexical bias.

These observations suggest that EEG2TEXT-CN is able to learn basic lexical patterns and limited semantic associations even in a zero-shot setting without visual cues. However, the model still faces difficulties in generating longer, coherent sentences. The frequent repetition of common words in predictions also indicates that the model favors high-probability tokens rather than generating diverse outputs.

Despite the modest BLEU scores, they show a positive correlation with qualitative alignment quality, confirming that BLEU can serve as a reasonable baseline metric for assessing EEG-to-text generation. In future work, metrics such as BERTScore or embedding-based cosine similarity may offer a more nuanced understanding of semantic alignment, especially in Chinese.

\section{Ablation Study: Impact of the Decoder}

To investigate the role of the decoder in our EEG2TEXT-CN architecture, we performed an ablation study by removing the transformer-based decoder and training an encoder-only variant. In this setup, the EEG encoder generates a sequence of embeddings from EEG input, and each embedding is directly mapped to a predicted character via a linear classification head without any autoregressive generation.

Formally, given input EEG segments \( \mathbf{X} = [\mathbf{X}_1, \dots, \mathbf{X}_N] \in \mathbb{R}^{N \times T \times C} \), the encoder produces feature vectors \( \mathbf{H} = [\mathbf{h}_1, \dots, \mathbf{h}_N] \in \mathbb{R}^{N \times d} \), which are passed through a shared linear classifier:
\[
\hat{y}_i = \text{Softmax}(W_c \cdot \mathbf{h}_i + b_c),
\]
where \( W_c \in \mathbb{R}^{|V| \times d} \) is the weight matrix and \( |V| \) is the vocabulary size. Unlike the full model, this encoder-only baseline lacks inter-token dependency modeling and language generation capabilities.

As shown in Table~\ref{tab:bleu_ablation} and the detail is in Table~\ref{tab:gt_pr_bleu_encoderonly}, the encoder-only model performs significantly worse than the full encoder-decoder model, especially in higher-order BLEU scores. This demonstrates the critical role of the decoder in capturing sequential and linguistic dependencies in EEG-to-text generation.

\begin{table}[h]
\centering
\caption{BLEU scores of the encoder-only model (ablation study).}
\label{tab:bleu_ablation}
\begin{tabular}{lcccc}
\toprule
\textbf{Model} & \textbf{BLEU-1} & \textbf{BLEU-2} & \textbf{BLEU-3} & \textbf{BLEU-4} \\
\midrule
EncoderOnly & 0.0674 & 0.0199 & 0.0139 & 0.0120 \\
\bottomrule
\end{tabular}
\end{table}

\begin{table}[h]
\centering
\caption{Comparison between Ground Truth and Predicted Results with BLEU scores (Encoder-Only Model)}
\label{tab:gt_pr_bleu_encoderonly}
\begin{tabular}{p{4.5cm} p{4.5cm} cccc}
\toprule
\textbf{Ground Truth (GT)} & \textbf{Predicted Result (PR)} & \textbf{BLEU-1} & \textbf{BLEU-2} & \textbf{BLEU-3} & \textbf{BLEU-4} \\
\midrule
\begin{CJK}{UTF8}{bsmi}1\end{CJK} & \begin{CJK}{UTF8}{bsmi}的\end{CJK} & 0.0000 & 0.0000 & 0.0000 & 0.0000 \\
\begin{CJK}{UTF8}{bsmi}全世界的狼都有一个共\end{CJK} & \begin{CJK}{UTF8}{bsmi}的的的的的的的的的的\end{CJK} & 0.1000 & 0.0333 & 0.0240 & 0.0211 \\
\begin{CJK}{UTF8}{bsmi}同的习性，在严寒的冬天\end{CJK} & \begin{CJK}{UTF8}{bsmi}的的的的的的的的的的\end{CJK} & 0.1810 & 0.0427 & 0.0274 & 0.0227 \\
\begin{CJK}{UTF8}{bsmi}集合成群，平时单身独处。\end{CJK} & \begin{CJK}{UTF8}{bsmi}的的的的的的的的的的\end{CJK} & 0.0000 & 0.0000 & 0.0000 & 0.0000 \\
\begin{CJK}{UTF8}{bsmi}眼下正是桃红柳绿的春\end{CJK} & \begin{CJK}{UTF8}{bsmi}的的的的的的的的的的\end{CJK} & 0.1000 & 0.0333 & 0.0240 & 0.0211 \\
\begin{CJK}{UTF8}{bsmi}天，日曲卡雪山的狼群按\end{CJK} & \begin{CJK}{UTF8}{bsmi}的的的的的的的的的的\end{CJK} & 0.0905 & 0.0302 & 0.0218 & 0.0191 \\
\begin{CJK}{UTF8}{bsmi}自然属性解体了，\end{CJK} & \begin{CJK}{UTF8}{bsmi}的的的的的的的\end{CJK} & 0.0000 & 0.0000 & 0.0000 & 0.0000 \\
\bottomrule
\end{tabular}
\end{table}


\section{Conclusion}
In this work, we introduce EEG2TEXT-CN, the first open-vocabulary EEG-to-Chinese generation model. Built upon a multiview EEG encoder and pretrained language model alignment via contrastive learning, our system operates under zero-shot conditions and successfully predicts full Chinese sentences from raw EEG data without access to visual stimuli or task-specific fine-tuning.

The model was trained and evaluated on the ChineseEEG dataset, where each EEG sequence corresponds to a subtitle-level Chinese sentence. While the absolute BLEU scores remain low, the model demonstrates the capacity to align partial lexical content with neural signals and marks a significant step forward in non-phonetic EEG-to-language research.

Future directions include: (1) improving long-range context modeling and memory; (2) incorporating semantic-aware evaluation metrics and human judgments; and (3) expanding the dataset with more participants and longer texts. EEG2TEXT-CN establishes a promising foundation for multilingual neural decoding and offers a new path for brain-to-language generation in underexplored linguistic domains.

\bibliographystyle{unsrt}
\bibliography{ref}

\end{document}